\documentclass[letterpaper, 10 pt, conference]{ieeeconf}  

\IEEEoverridecommandlockouts                              

\usepackage{multirow}
\usepackage{booktabs}
\usepackage{subcaption}
\usepackage{hyperref}
\usepackage{graphics} 
\usepackage{epsfig} 
\usepackage{mathptmx} 
\usepackage{times} 
\usepackage{amsmath} 
\usepackage{amssymb}  
\hypersetup{
    colorlinks=true,     
    linkcolor=black,     
    urlcolor=black,      
    citecolor=black,     
    filecolor=black,     
    pdfborder={0 0 0}    
}
\graphicspath{{./figures/}}
\title{\LARGE \bf
Visual Deformation Detection Using Soft Material Simulation for Pre-training of Condition Assessment Models
}

\author{Joel Sol$^{1}$,  Amir M. Soufi Enayati$^{2}$ and Homayoun Najjaran$^{3}$
\thanks{$^{1}$ $^{2}$ $^{3}$ Joel Sol, Amir M. Soufi Enayati and Homayoun Najjaran are with  Faculty of Electrical and Computer Engineering,
        University of Victoria, Victoria, BC V8P 5C2, Canada
        {\tt\small \{joelsol, amsoufi, najjaran\}@uvic.ca}}%
\thanks{The authors have provided supplementary material (code) available at \url{https://github.com/JoelESol/Synthetic_Deformation_Data_Generation}}
}

\begin{document}

\maketitle
\thispagestyle{empty}
\pagestyle{empty}

\begin{abstract}    
    This paper addresses the challenge of geometric quality assurance in manufacturing, particularly when human assessment is required. It proposes using Blender, an open-source simulation tool, to create synthetic datasets for machine learning (ML) models. The process involves translating expert information into shape key parameters to simulate deformations, generating images for both deformed and non-deformed objects. The study explores the impact of discrepancies between real and simulated environments on ML model performance and investigates the effect of different simulation backgrounds on model sensitivity. Additionally, the study aims to enhance the model's robustness to camera positioning by generating datasets with a variety of randomized viewpoints. The entire process, from data synthesis to model training and testing, is implemented using a Python API interfacing with Blender. An experiment with a soda can object validates the accuracy of the proposed pipeline.
\end{abstract}

\section{Introduction}
The process of geometrical quality assurance based on the coordinate measurement is an essential aspect of ensuring the accuracy and consistency of manufactured products in assembly lines. By utilizing this method, manufacturers can reduce production time and costs, while also increasing the overall quality of their precision products~\cite{pyzdek2003quality}. The accuracy of geometrical quality inspection in industrial settings heavily relies on the visual or haptic assessment of human operators, which can be time-consuming and prone to errors~\cite{duan2020real}. This can lead to significant delays and increased costs in the manufacturing process, making it a critical issue that needs to be addressed.

Heuristic methods, such as conventional photogrammetry~\cite{ben2019automatic}, have been popular in industrial inspection, but they often lack generalizability and are limited by the complexity of the inspection task. As a result, there is a growing interest in the development of more advanced and automated inspection methods, such as computer vision and machine learning, that can provide more accurate and efficient solutions to quality assurance in industrial settings. Modern quality assurance methods powered by machine learning can solve this problem~\cite{godina2019quality}, but require large datasets with diverse deformations for training~\cite{talha2022}. Generating a large and diverse dataset of deformations in physical objects can be a difficult and time-consuming task. Moreover, the physical limitations and constraints in object deformations might not cover all the possible scenarios, making it difficult to achieve a comprehensive dataset~\cite{boikov2021synthetic}.

Therefore, alternative approaches, such as simulations, can provide a more efficient and versatile way of generating such datasets for machine learning-based quality assurance methods~\cite{gutierrez2021synthetic}. While simulation is a viable option for creating training datasets, there have been efforts to explore data augmentation~\cite{tsirikoglou2020survey} or generation methods like generative adversarial networks (GAN)~\cite{saiz2021generative}. However, these methods are not as interpretable or controllable by expert knowledge, and may not provide the same level of meaningful diversity as simulations. Independent research has been conducted in this area, specifically aimed at automating the simulation of deformations~\cite{wakamatsu2005dynamic}. As a result, this technology is now available for data synthesis applications. This advancement has enhanced the efficiency and accuracy of data synthesis, enabling researchers to generate realistic simulations of deformations more quickly and reliably.

While simulated input can be useful in training machine learning models, there is a risk that the model's performance will be degraded when transferred to a physical environment due to discrepancies with the real world. Therefore, it is important to consider the limitations of simulated input. One approach to improving the robustness of machine learning models in quality control applications is to add noise to images~\cite{zheng2016improving}. This can help the model to better generalize and perform well on unseen data in the real world. However, it is important to carefully select the type and amount of noise to add, as too much noise can spoil the model's performance. Another factor to consider when working with machine vision is the camera position and calibration. In order to ensure faster and more accurate evaluations in the real world, it is imperative to reduce the sensitivity of machine vision to camera calibration~\cite{barazzetti2012targetless}.

Geometrical assessment of manufacturing products has taken advantage of a wide range of measurement techniques to accurately identify and classify deformities in objects. Researchers have explored different methods for shape measurement, including the use of laser optics~\cite{RAMESHKUMAR20222265, YAN2022169923}. However, this technique can be expensive and requires specialized equipment, making it less accessible for some applications. An alternative approach that has gained popularity in recent years is the use of RGB-Depth (RGB-D) images for deformity detection and classification. These images capture both the color and depth information of objects, allowing for a more comprehensive analysis of their geometries. By leveraging machine learning algorithms and computer vision techniques, researchers have been able to identify different types of deformations in objects, such as rigid, elastic, plastic, etc., from images~\cite{robotics6010005}.

On one hand, The advantage of using RGB-D images over laser optics is the ease and accessibility of data collection which allows for faster and more cost-effective data acquisition, making it a viable option for many industrial and manufacturing settings. On the other hand, using point-cloud laser optic measurement of objects in combination with image inputs has shown great accuracy suitable for real-time deformation detection in an industrial setting~\cite{TANG201936}. Another approach in this regard is using conventional heuristic photogrammetry techniques to obtain 2D readings from 3D CAD models and compare the outcome with real 2D images taken from the object~\cite{ben2019automatic}. Overall, our concern in this paper would be the availability and easier implementation to be able to address a wider range of applications. Therefore, exploiting raw CAD files and RGB images would be the priority over using other techniques.

Machine learning manufacturing applications using image analysis have been a research topic for many years in the manufacturing and production industries~\cite{staar2019anomaly}. For instance, in a recent paper, a computer vision module was designed to substitute human judgment in wear analysis as a part of manufacturing assessments~\cite{Xiao2023}. Additionally, supervised learning although the predominant machine learning framework, is not the only one explored. A semi-supervised learning pipeline suggested by~\cite{liu2021semi}, uses Auto-encoders to reduce the dependence of regular supervised-learning-based methods on huge labeled datasets. Although this improvement will facilitate the inclusion of unlabeled data in many cases, the total available data, whether labeled or not, might not always be enough. This necessitates using simulation-based data generation pipelines. The competence of such an approach has been verified in different problems of this field~\cite{robotpose_etfa2019_cheind}, and in this paper, we proposed a pipeline for another application whose efficiency is validated in a simulated environment.

\begin{figure}[thpb]
\centerline{\includegraphics[width=0.4\textwidth]{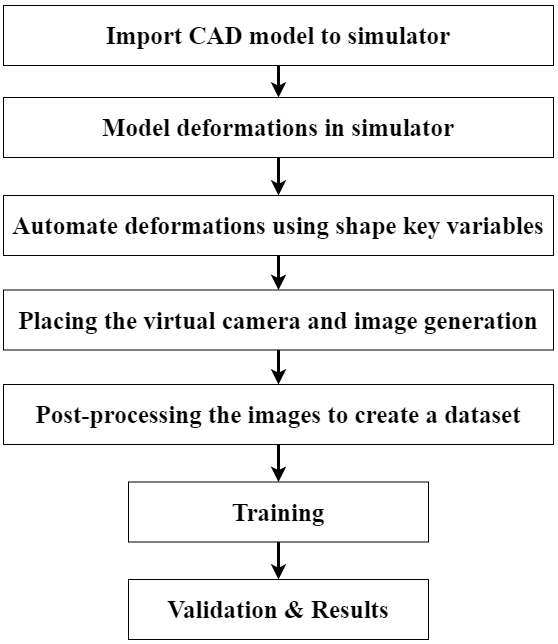}}
\caption{Proposed simulation-based deformation inspection pipeline.}
\label{fig:scheme}
\end{figure}

We make use of an open-source graphics simulation tool, to automate the creation of synthetic datasets for machine learning (ML) models. The graphics engine uses shape key parameters to simulate object deformities and generate datasets for classification models. Randomized camera viewpoints are used to reduce the need for labor-intensive camera calibration in industrial settings. An experiment is conducted to assess the performance of the proposed pipeline, and the results show promise for the sim-to-real transfer of ML models trained using this method. The proposed scheme is illustrated in Figure \ref{fig:scheme}. Its entire data synthesis, model training, and testing stages have been implemented using an integrated computer script.

The paper is organized as the following. First, we introduce the environment chosen for showcasing our proposed method in Section \ref{sec:env}. Next, we describe the data generation procedure in Section \ref{sec:data}. We introduce the machine learning model architecture and training process in Section \ref{sec:arch}. Finally, we conclude the paper with the discussion of results and conclusion in Sections \ref{sec:result} and \ref{sec:conclusions}.

\section{Environment}\label{sec:env}
The environment used for the simulation is Blender. Blender is an open-source computer graphics, rendering, and simulation environment. Blender accepts most file types associated with computer-aided design (CAD) and is capable of animating deformations and defects. The Python interface allows for the automation of procedural deformations and defects, and various camera positions for creating a dataset from a CAD design. Blendtorch is a Python framework for integrating Blender into Pytorch for deep learning applications \cite{blendtorch_icpr2020_cheind}. Blendtorch was used to integrate the Blender Python scripting environment into an integrated development environment (IDE) with the rest of the project so rendered images could be used to automate dataset generation. A soda can was used to model deformation and dataset creation. The object was selected as it is easy to acquire a sizeable physical dataset for and easy to deform to create the deformed class. The metallic nature of the object also presents an interesting classification challenge. 

\begin{table}[htbp]
\centering
\caption{Camera position parameters in Blender}\label{tab1}
\begin{tabular}{@{}ccccc@{}}
\multicolumn{1}{l}{\multirow{2}{*}{}}
& \multicolumn{4}{c}{Camera} \\ \cmidrule(l){2-5} 
\multicolumn{1}{l}{} & No. 1 & No. 2 & No.3 & No.4 \\ \midrule
$\theta$ & $[20^{\circ},70^{\circ}]$ & $[110^{\circ},160^{\circ}]$ & $[200^{\circ},250^{\circ}]$ & $[290^{\circ},340^{\circ}]$ \\ \midrule
$\phi$ & \multicolumn{4}{c}{$[50^{\circ},70^{\circ}]$ for all cameras} \\ \midrule
$r$ & \multicolumn{4}{c}{$[0.3m-0.45m]$ for all cameras} \\ \bottomrule
\end{tabular}
\end{table}

Deformation of the object was done by combining a lattice deformation and a displace modifier. A simple cube lattice was created around the can to control major deformations coined as \textit{crushes}, \textit{pinches}, \textit{folds}, \textit{twists}, and \textit{crunches}. When the lattice is deformed it controls the resulting deformation in the soda can. Twelve lattice deform shape keys were created of the various macro deformation types. Movement and rotation deformations were applied to both the seal and tab of the soda cans and mapped to shape keys and make the cans appear open.
Smaller deformations on the surface of the can were created by using a displace modifier. Vertex groups were used to limit these deformations to the side of the can leaving the top and bottom unaffected. The displace modifier slightly inflates the object and applies a texture to deform the surface. For the soda can object, the most realistic texture was a hard \textit{Stucci} texture scaled to look similar to the sharp edges of a crinkled pop can. Three different hard \textit{Stucci} textures were created and assigned to a corresponding shape key.

By combining the lattice deformation and displacement deformations, high quality procedural deformations can be created. The shading of the environment was set up to create a realistic looking can, with a UV mapped label applied to the side and an aluminum like metallic texture applied to the top and bottom.

A dynamic procedural lighting environment was created using the Blender world shading module. An EXR file was randomly selected and given a random rotation to create realistic and different light and shadows on the can object. The background of the object was either set to black or green based on if the lighting would hit the object from the camera view. The cameras were placed in a uniform random polar position as shown in Table~\ref{tab1}. The images were rendered and saved as 512x512 RGB images. Later in the pipeline, these images were modified.

\section{Dataset Construction and Post-Processing}\label{sec:data}

\begin{figure*}[htbp]
\centering
     \begin{subfigure}[b]{0.40\textwidth}
         \centering
         \includegraphics[width=0.98\textwidth]{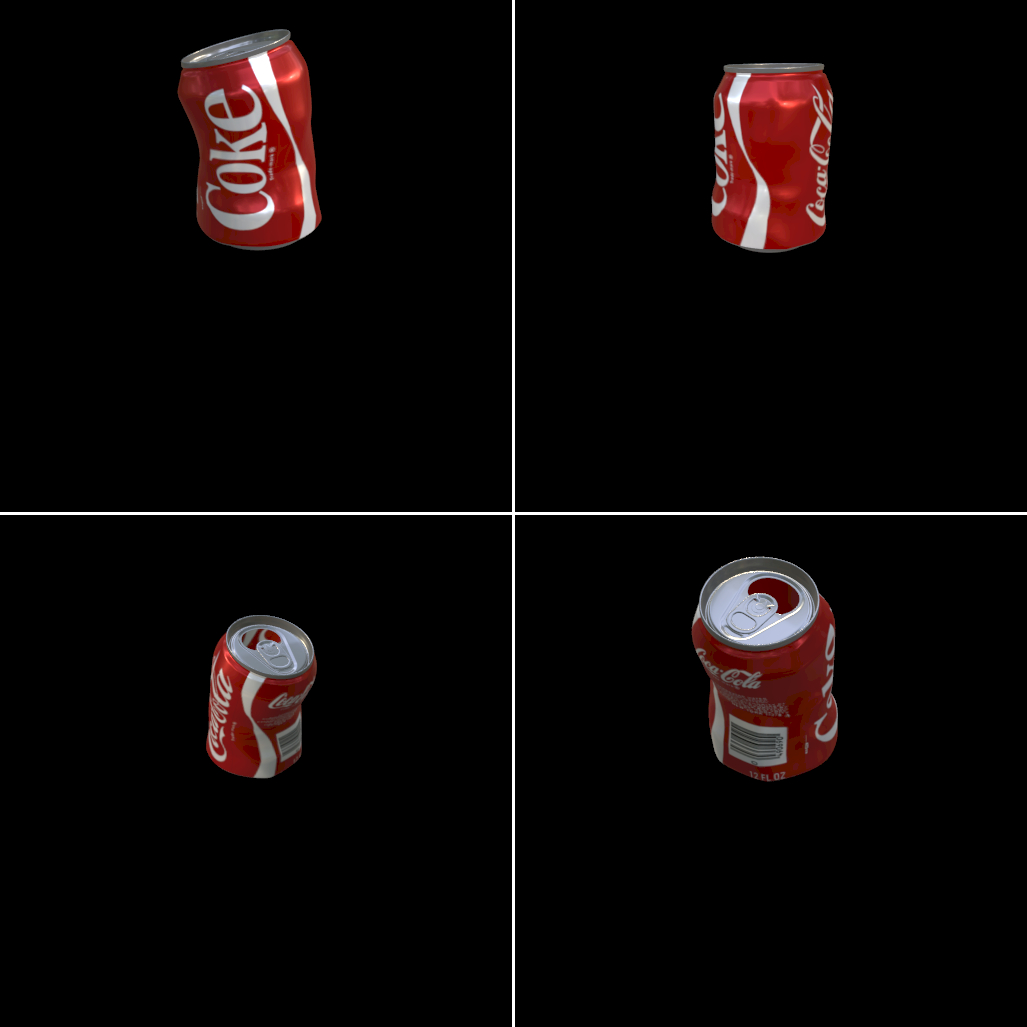}
         \caption{Synthetic deformed can black background}
         \label{fig1-1}
     \end{subfigure}
     \hspace{0.1cm}
     \begin{subfigure}[b]{0.40\textwidth}
         \centering
         \includegraphics[width=0.98\textwidth]{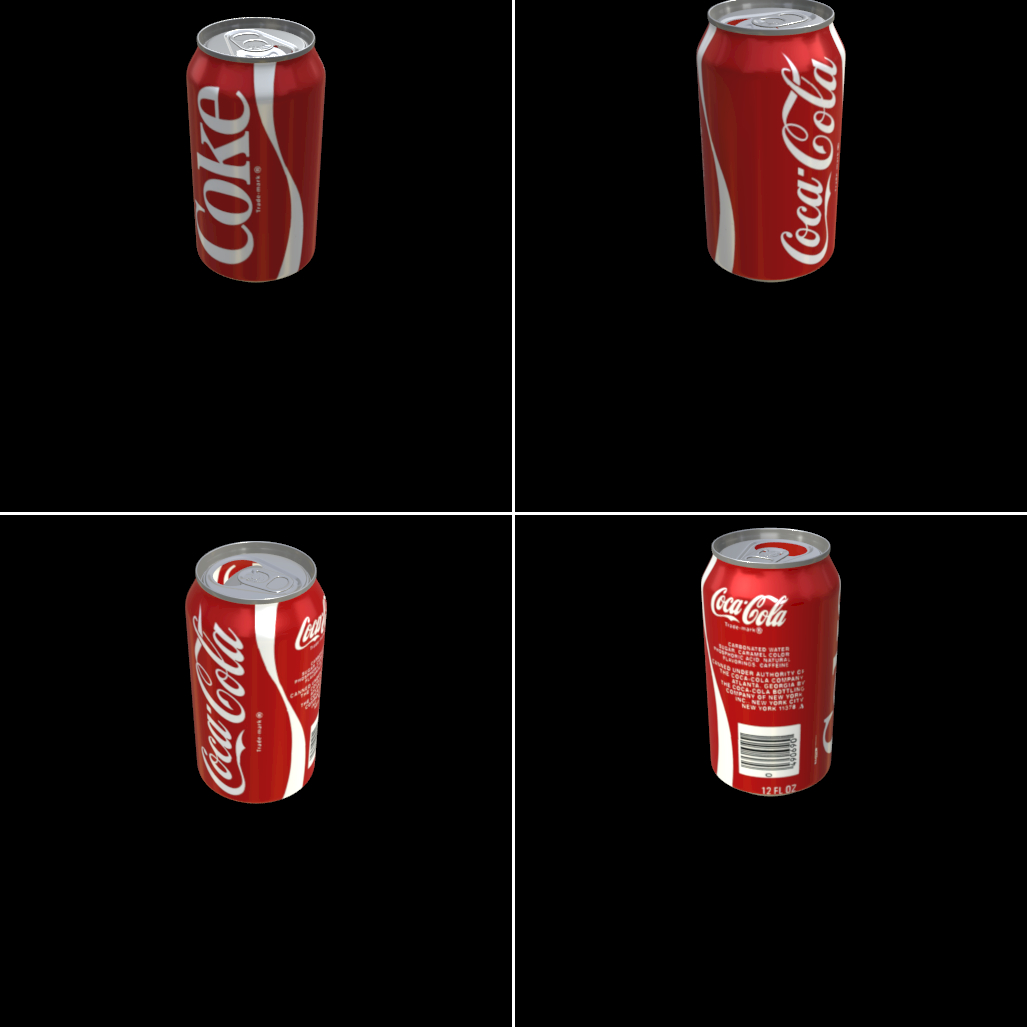}
         \caption{Synthetic non-deformed can black background}
         \label{fig1-2}
     \end{subfigure}
     \hspace{0.1cm}
     \begin{subfigure}[b]{0.40\textwidth}
         \centering
         \includegraphics[width=0.98\textwidth]{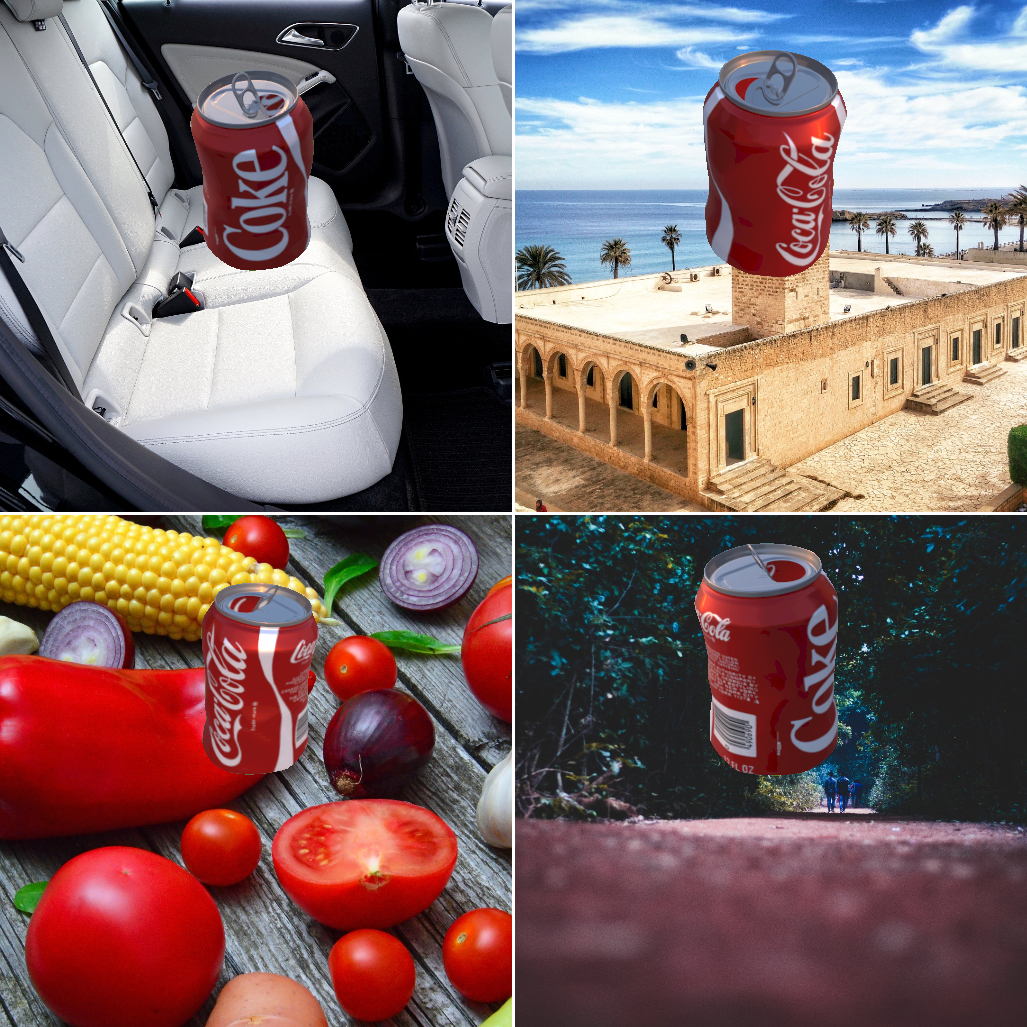}
         \caption{Synthetic deformed can BG-20k background}
         \label{fig1-3}
     \end{subfigure}
     \hspace{0.1cm}
     \begin{subfigure}[b]{0.40\textwidth}
         \centering
         \includegraphics[width=0.98\textwidth]{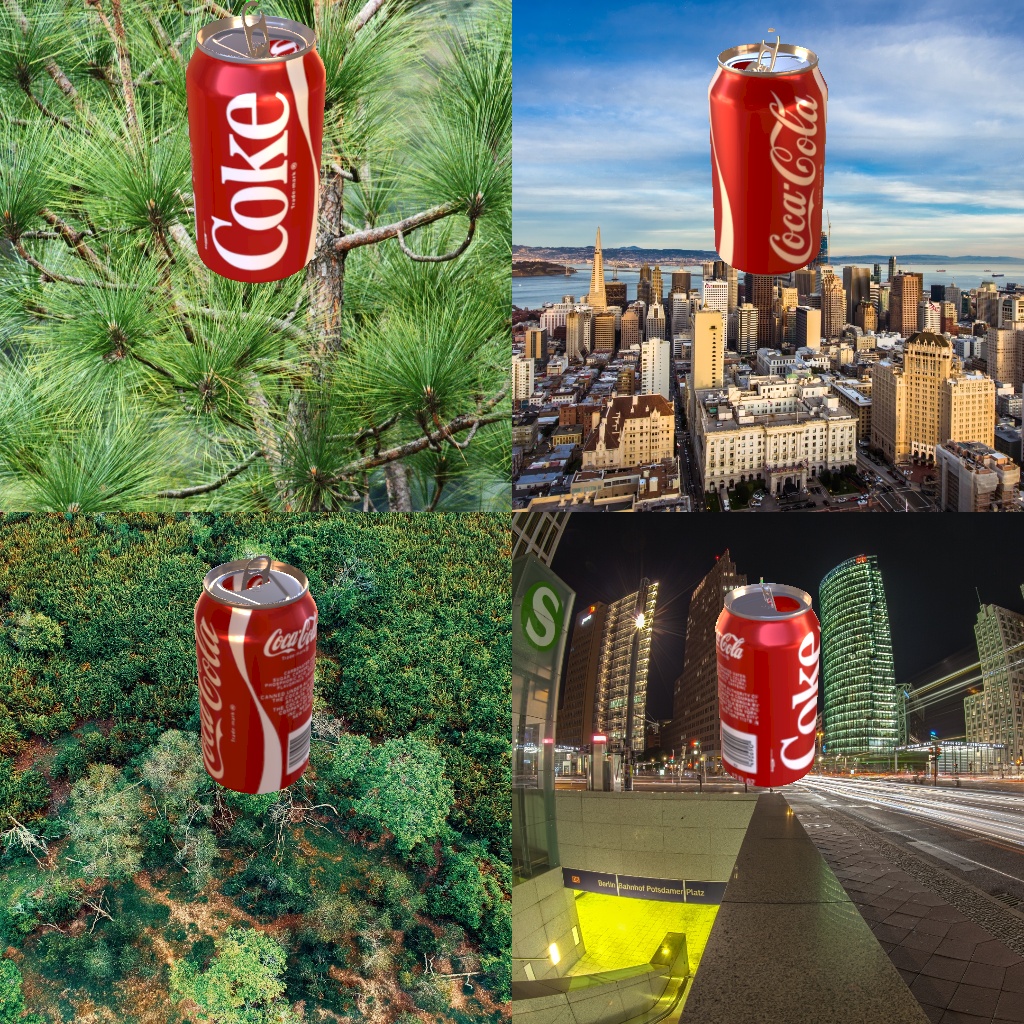}
         \caption{Synthetic non-deformed can BG-20k background}
         \label{fig1-4}
     \end{subfigure}
     \hspace{0.1cm}
     \begin{subfigure}[b]{0.40\textwidth}
         \centering
         \includegraphics[width=0.98\textwidth]{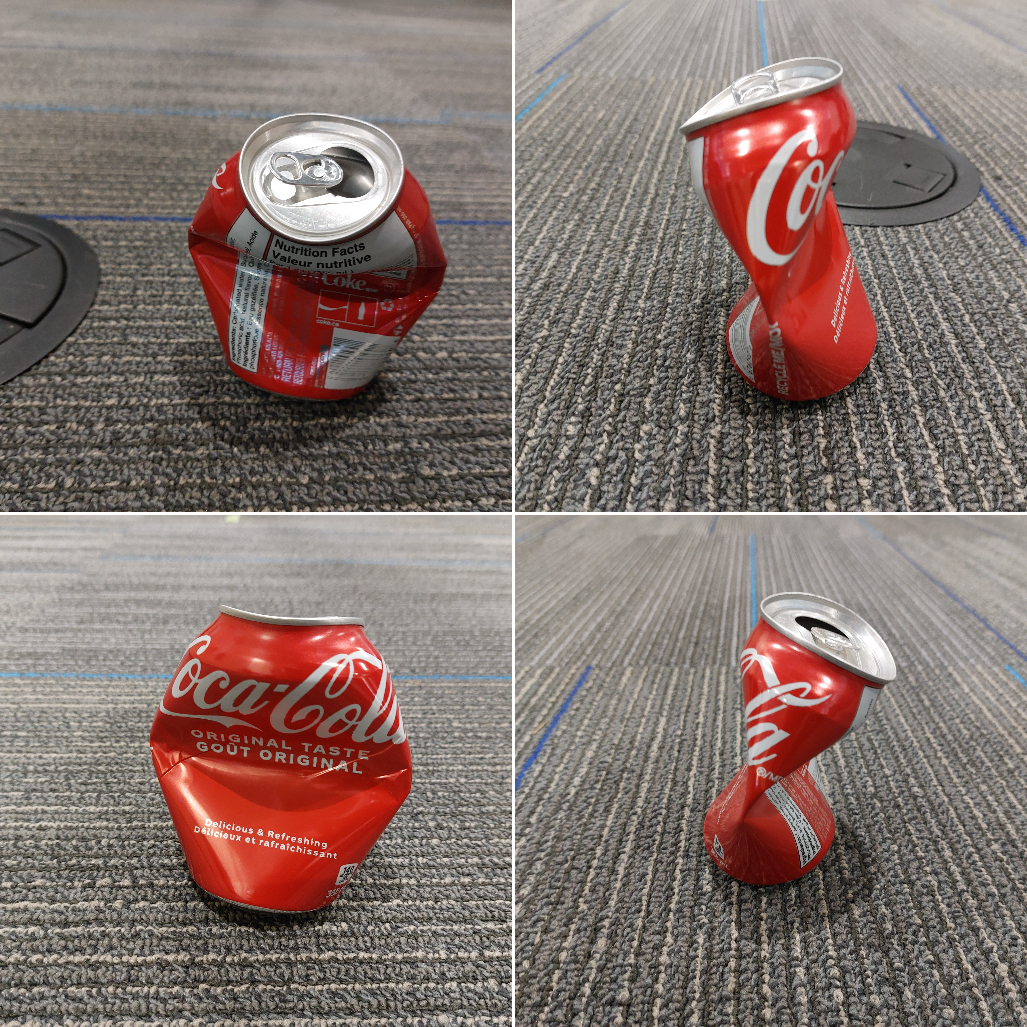}
         \caption{Real deformed can}
         \label{fig1-5}
     \end{subfigure}
     \hspace{0.1cm}
     \begin{subfigure}[b]{0.40\textwidth}
         \centering
         \includegraphics[width=0.98\textwidth]{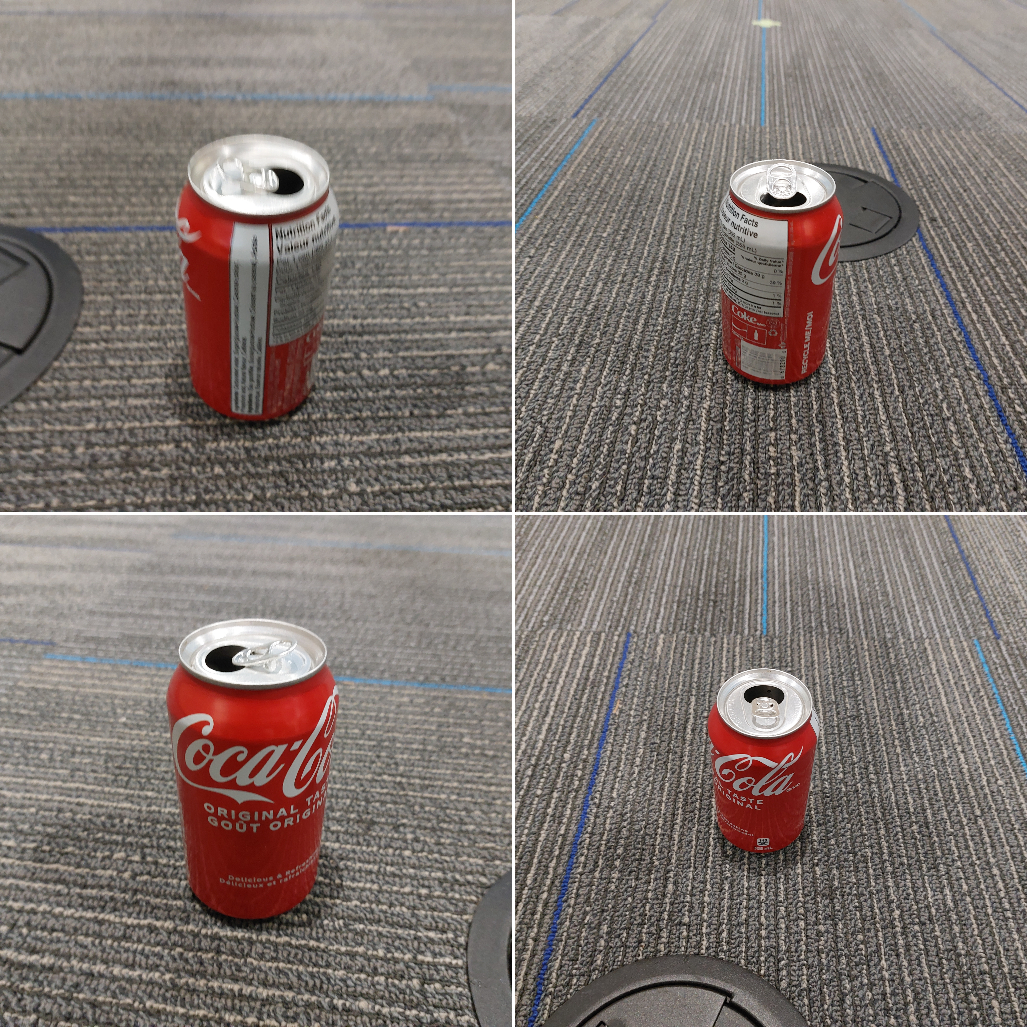}
         \caption{Real non-deformed can}
         \label{fig1-6}
     \end{subfigure}
     \hspace{0.1cm}
    \caption{Synthetic and real can dataset examples}
    \label{fig1}
\end{figure*}

Images rendered in the Blender environment were used to create a dataset. A total of 6000 images of the cans were used, 3000 of which were of the non-deformed object and 3000 were deformed. In all virtual samples, the tab and seal of the soda cans are set to a random state between fully open and fully closed. The tab is always set to a state more closed than the seal to prevent impossible geometries and part clipping. Dataset samples of the non-deformed object is shown in Figure \ref{fig1-2}.

In the deformed class, 12 lattice deformations and 3 displacement deformations were combined. The displacement deformations were each assigned a randomized weight and combined in a way to ensure there was a procedural deformation on the surface of the object. For the lattice deformations, three or more types (\textit{crush}, \textit{pinch}, etc.) were randomly selected and combined. Each type was assigned a randomized weight to differ the extent each lattice deformation type makes on the object. For each lattice deformation type, various shape keys were mixed and selected. The end result is a realistic, procedural deformation. Examples of the deformed class are shown in Figure \ref{fig1-1}. These 6000 images were used to create the synthetic dataset.

The synthetic dataset was modified prior to training to facilitate model generalization. The background was initially rendered to an RGB value of (0, 255, 0) to create a green screen. Morphological operations such as erosion, opening and closing were used to create a high quality mask for background transfer. These operations are essential to prevent any holes that may be accidentally created in the mask and ensure no green pixels are left over which could affect training. The background images for the synthetic dataset were randomly selected from the BG-20k dataset\cite{bg20k}. This dataset was chosen as it was created for image matting tasks where foreground features are recognized from a background image. This is ideal for the use case of a generalized background with the important can features in the foreground. Examples of these synthetic images are shown in Figure \ref{fig1-3} and Figure \ref{fig1-4} for the deformed and non-deformed classes.

A physical dataset was created as well in order to provide a real-world object to test our synthetic data against and measure the sim-to-real performance gap. Like the synthetic dataset, 4 camera views of each deformed object were used. To create the 160 images making up the deformed class, 40 deformed cans were used. Three cans were used for the non-deformed class with five images from each camera view quadrant. These 60 images are used for the non-deformed class for the physical dataset. The images were taken with a smart phone camera in a 1:1 aspect ratio and in positions and angles similar to those set up in the synthetic dataset. Examples for the real-world data are shown in Figure \ref{fig1-5} and Figure \ref{fig1-6}.

\section{Network and Training}\label{sec:arch}

We use a modified version of pre-trained VGG-16 to analyze the sim-to-real crossing. The VGG-16 model architecture, proposed by Simonyan and Zisserman\cite{VGG-16}, has been shown to be widely useful in various computer vision tasks such as image classification, object detection, and image segmentation. The simplicity of the architecture makes it easily adaptable and effective, making it a good network to use as a benchmark. The original VGG-16 model consists of 13 convolutional layers and 3 fully connected layers. The convolutional layers use 3-by-3 filters with a stride and padding of 1 to maintain the spatial dimensions of the input. Max pooling layers use a 2-by-2 filter with a stride of 2 and are used after every 2 or 3 convolution layers. These design characteristics reduce the spatial dimensions by half.

The VGG-16 network explained above, has been modified to adapt to our specific task. The network has the same convolution and max pooling layers, but the fully connected layers have been changed. After the convolution layers, an adaptive average pooling layer is used to fix the output size. This is followed by 3 fully connected layers with ReLU activation and a dropout layer. The 3 fully connected layers go to a one-hot output with a sigmoid activation function. The sigmoid activation function has been shown to be the optimal activation function in feed-forward binary classification networks \cite{https://doi.org/10.48550/arxiv.1811.03378}.

While training, all convolution feature layers except the final layer were frozen in order to preserve learned features from previous training on ImageNet-1k. This pre-training allows the model to learn the general features of images, which can be transferred to the task of classifying the deformity state of cans. For both the black background and BG-20k backgrounds, the synthetic dataset was split into training and testing data and evaluated. For the sim-to-real training, the entire synthetic dataset was used for training and the real-world dataset was used for testing. The network was trained using the Adam optimizer with a learning rate of 0.001 and a batch size of 32, for 15 epochs.

\section{Results}\label{sec:result}

\begin{figure}[htbp]
\centering
     \begin{subfigure}[b]{0.23\textwidth}
         \centering
         \includegraphics[width=0.98\textwidth]{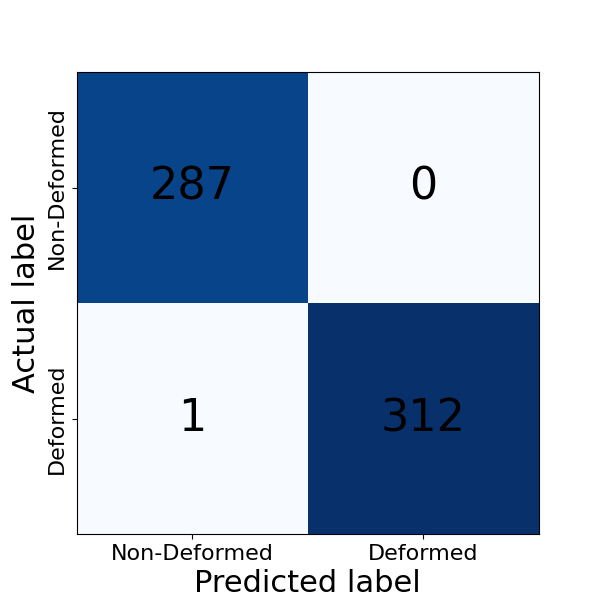}
         \caption{Black background - synthetic dataset}
         \label{fig:confusion_1}
     \end{subfigure}
     \begin{subfigure}[b]{0.23\textwidth}
         \centering
         \includegraphics[width=0.98\textwidth]{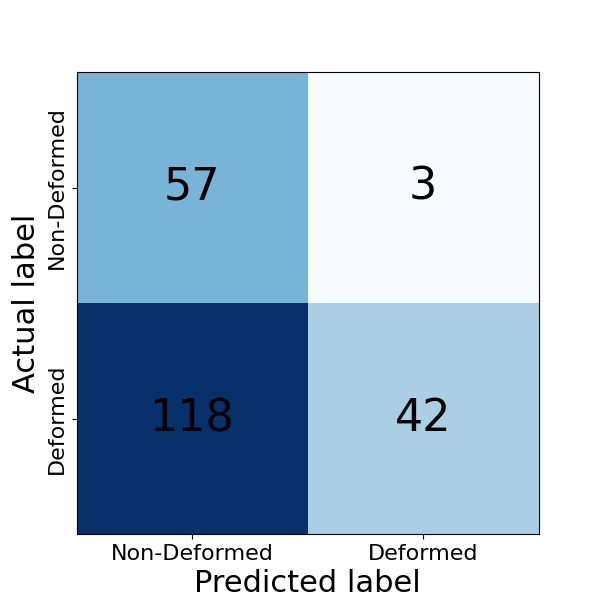}
         \caption{Black background - sim-to-real}
         \label{fig:confusion_2}
     \end{subfigure}
     \begin{subfigure}[b]{0.23\textwidth}
         \centering
         \includegraphics[width=0.98\textwidth]{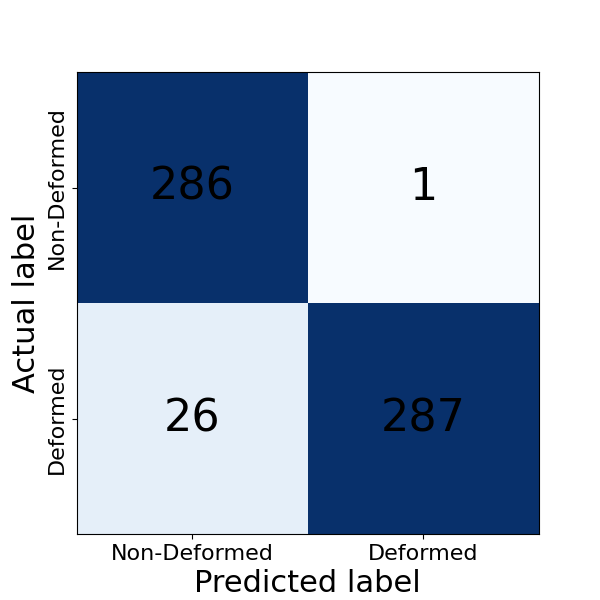}
         \caption{BG-20k background - synthetic dataset}
         \label{fig:confusion_3}
     \end{subfigure}
     \begin{subfigure}[b]{0.23\textwidth}
         \centering
         \includegraphics[width=0.98\textwidth]{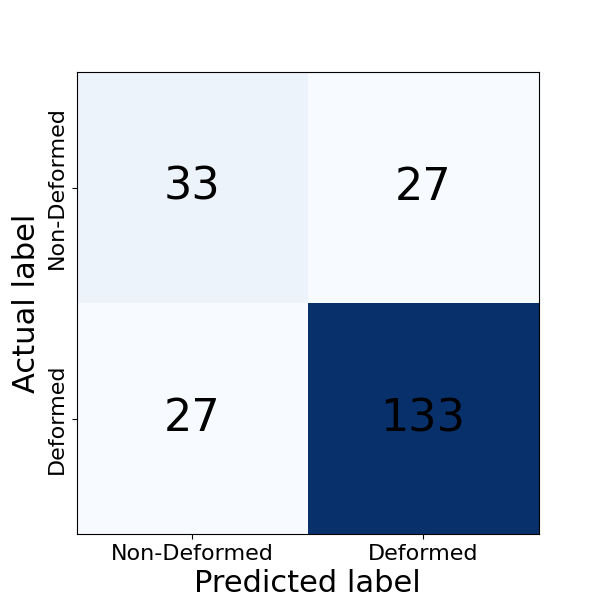}
         \caption{BG-20k background - sim-to-real}
         \label{fig:confusion_4}
     \end{subfigure}
    \caption{Confusion matrices for the network performance}
    \label{fig:confusion}
\end{figure}

The results for the network's training can be found in Table \ref{tab2} and Figure \ref{fig:confusion}. Overall, VGG-16 was able to accurately detect deformation in both synthetic datasets but saw a significant decrease when validated on real-world data. This is expected, due to different deformations, camera angles, camera intrinsics, and lighting. Using BG-20k as a background saw a smaller sim-to-real gap than just a black background and therefore a significant improvement in generalization. 

The black background dataset had the best synthetic accuracy as shown in Figure \ref{fig:confusion_1} and had almost perfect sim-to-real accuracy. However, the results suffered significantly trying to cross the sim-to-real gap as shown in Figure \ref{fig:confusion_2}.

The BG-20k background dataset had great synthetic accuracy but was slightly worse than the black background dataset as shown in Figure \ref{fig:confusion_3}. The sim-to-real results still suffered an accuracy penalty but were much improved in comparison to the black background dataset as shown in Figure \ref{fig:confusion_4}.

\begin{table}[pbht]
\centering
\caption{Performance metrics of VGG-16}\label{tab2}
\begin{tabular}{lcc|cc}
\multirow{2}{*}{Metrics} & \multicolumn{2}{c|}{Synthetic dataset} & \multicolumn{2}{c}{Sim-to-real} \\ \cline{2-5}
 & Black & BG-20k & Black & BG-20k \\ \hline
{$Accuracy$\quad} & 0.998 & 0.955 & 0.450 & 0.755 \\ \hline
{$F_1$\quad} & 0.998 & 0.955 & 0.485 & 0.550 \\ \hline
{$Recall$\quad} & 1.0 & 0.997 & 0.950 & 0.550 \\ \hline
{$Precision$\quad} & 0.997 & 0.917 & 0.326 & 0.550 \\ \hline
\end{tabular}
\end{table}

\begin{figure}[htbp]
    \centering
    \includegraphics[width=0.50\textwidth]{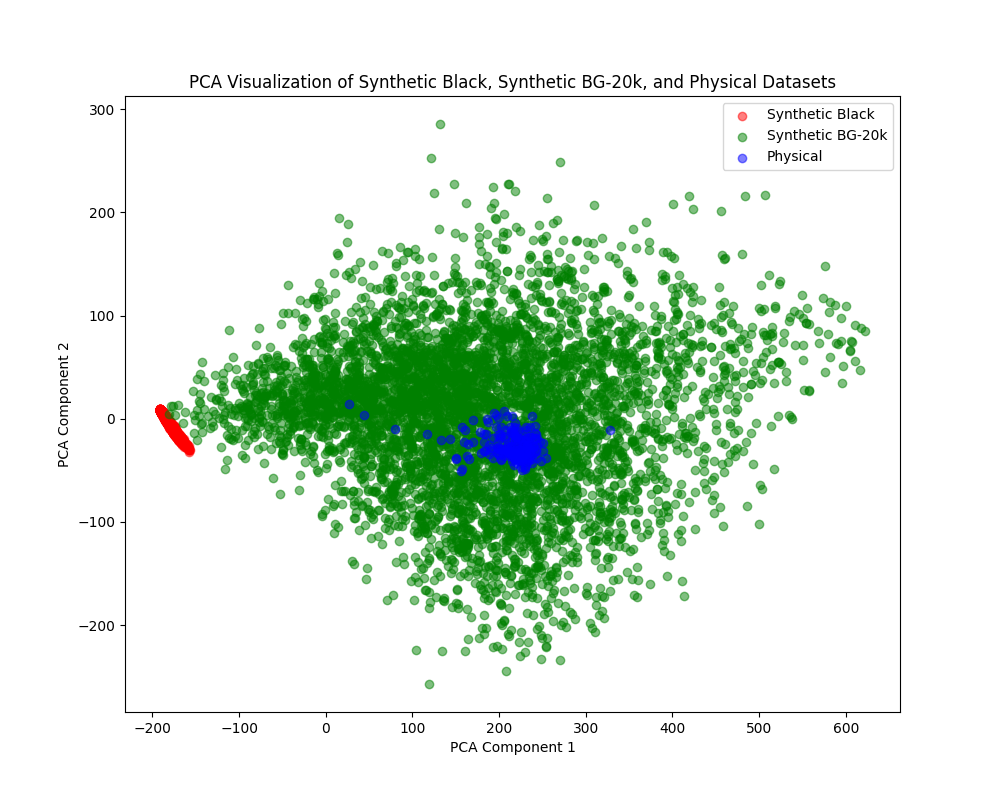}
    \caption{PCA visualization of all datasets}
    \label{fig:PCA}
\end{figure}

The PCA visualization shown in Figure \ref{fig:PCA} showcases the distributions of the datasets. Adding in the BG-20k background causes greater variation and more significant overlap than just the black background. This results in the network trained on synthetic BG-20k dataset to be better at predicting real-world deformations. The distribution of the black background is very tightly spaced with a significant difference to the physical dataset. The resulting performance increase from this more varied distribution is showcased in Table \ref{tab2}.

\section{Discussion}
There are several implications due to the performance of this method. This provides a method for condition assessment and deformation detection for delicate objects, that require noninvasive methods, or where haptic sensing is unsuitable. Modern quality assurance is increasingly reliant on ML techniques which require extensive datasets. These datasets can be costly and time-consuming to create. Using the synthetic data created using this pipeline would reduce the cost and time required to create datasets.

As this method relies on the simulation quality, improvements to the simulation quality are expected to improve all sim-to-real metrics and real-world generalizability. Additional and more accurate shape keys could improve dataset quality due to more accurate simulation. Results are also expected to improve using depth cameras for the real-world dataset capture and compositing depth information in Blender. This is expected to improve results as well as allow for finer deformations to be detected.

A future direction for this work could be using transfer learning or network improvements. Transfer learning would allow for the synthetic data to more accurately resemble the real-world data to adapt the synthetic dataset to a specific real-world scenario instead of generalization. Due to the camera views being stochastic, using generative adversarial networks (GANs) for unsupervised domain adaptation is a suitable future direction. However, these methods would require a limited physical dataset for pre-training the condition assessment network. More experimentation with domain generalization network architectures is also another avenue for future direction.

\section{Conclusion}\label{sec:conclusions}
Quality assurance of objects requires expensive equipment and labor-intensive processes. In this work, we have proposed a novel pipeline for geometrical quality assurance that significantly reduces time and effort in industrial settings. The main contribution to the novelty of this pipeline is the use of Blender to facilitate data synthesis. Through expert knowledge, deformations can be applied to the virtual object negating the need for a physical dataset. Using the shape keys to apply various deformations in Blender, a large varietal dataset can be created for ML classification. The accuracy shown from both networks shows that the pipeline is effective at pre-training a network to detect deformations and cross the sim-to-real gap. This is particularly useful in cases where the object of quality control is difficult to create a dataset, e.g., the object is fragile, difficult to handle, expensive, and visual methods are required. The pipeline shown could be used to bring modern quality assurance to processes still reliant on human operators. The pipeline has shown good results and serves as an excellent solution or starting point for pre-training a condition assessment model for other objects. A significant strength of this method is that no real-world data is required to train the condition assessment model while having good performance metrics. Our ongoing research aims to extend the proposed pipeline to adapt the synthetic domain to a specific real-world domain using a generative adversarial network (GAN) as opposed to background generalization.

\addtolength{\textheight}{-12cm}   





{
\bibliographystyle{IEEEtran}
\bibliography{ref}
}

\end{document}